\pdfoutput=1

\documentclass[11pt]{article}

\usepackage[]{acl}

\usepackage{times}
\usepackage{latexsym}
\usepackage{adjustbox}
\usepackage{multicol}
\usepackage{multirow}
\usepackage{ctable}
\usepackage{arydshln}
\usepackage{amsmath}
\usepackage{hyperref}
\usepackage[T1]{fontenc}

\usepackage[utf8]{inputenc}

\usepackage{microtype}

\title{Improving Query-Focused Meeting Summarization with \\ Query-Relevant Knowledge}

\author{Tiezheng Yu , Ziwei Ji, Pascale Fung \\
Center for Artificial Intelligence Research (CAiRE)\\
Department of Electronic and Computer Engineering\\
The Hong Kong University of Science and Technology, Clear Water Bay, Hong Kong\\
\texttt{\{tyuah,zjiad\}@connect.ust.hk},  \texttt{pascale@ece.ust.hk}}

\begin{document}
\maketitle
\begin{abstract}
Query-Focused Meeting Summarization (QFMS) aims to generate a summary of a given meeting transcript conditioned upon a query. The main challenges for QFMS are the long input text length and sparse query-relevant information in the meeting transcript. In this paper, we propose a knowledge-enhanced two-stage framework called Knowledge-Aware Summarizer (KAS) to tackle the challenges. In the first stage, we introduce knowledge-aware scores  to improve the query-relevant segment extraction. In the second stage, we incorporate query-relevant knowledge in the summary generation. Experimental results on the QMSum dataset show that our approach achieves state-of-the-art performance. Further analysis proves the competency of our methods in generating relevant and faithful summaries.
\end{abstract}

\section{Introduction}
Meetings are an essential part of human collaboration and communication. Especially in recent years, the outbreak of Covid-19 has led people to meet online, where most meetings are automatically recorded and transcribed. Query-Focused Meeting Summarization (QFMS)~\cite{zhong2021qmsum} aims to summarize the given meeting transcript conditioned upon a query, which helps people efficiently catch up to the specific part of the meeting they want to know.

There are two main challenges for QFMS: Firstly, meeting transcripts can be so long that current deep learning models cannot encode them at once. Even for models~\cite{beltagy2020longformer,xiong2022adapting} that accept long text input, the cost of computational complexity is enormous. Secondly, query-relevant content is sparsely scattered in the meeting transcripts, meaning a significant part of the transcripts is noisy information when given a particular query. Therefore, the models need to reduce the noisy information's impact effectively.

In this paper, we focus on the two-stage framework, extracting query-relevant segments from the meeting transcripts and then generating the summary based on the selected content. Compared to the end-to-end approaches~\cite{zhu2020hierarchical,pagnoni2022socratic} that directly encode the entire meeting transcripts, the two-stage framework is better at keeping computational efficiency and easier to scale up to longer inputs. Specifically, we propose Knowledge-Aware Summarizer (KAS) that incorporates query-relevant knowledge in both stages. In the first stage, we extract knowledge triples from the text segments and introduce knowledge-aware scores to improve segment ranking. In the second stage, the extracted query-relevant knowledge triples are utilized as extra input for the summary generation. We conduct experiments on a QFMS dataset named QMSum~\cite{zhong2021qmsum} and achieve state-of-the-art performance. We further investigate how the different numbers of extracted segments affect the final performance. In addition, we manually evaluate the generation quality regarding fluency, relevance and factual correctness.

\begin{figure}[t]
    \centering
    \includegraphics[width=1\linewidth]{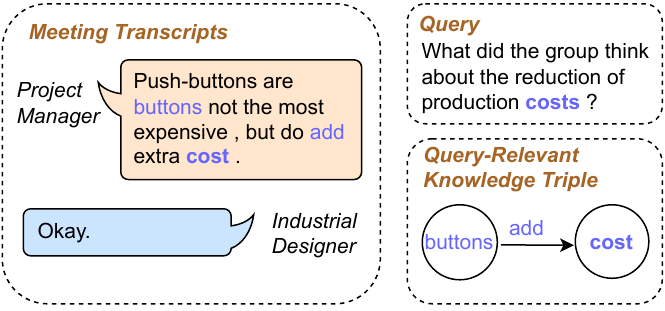}
    \caption{An example of query-relevant knowledge triple extracted from meeting transcripts. The knowledge triple can be used in query-relevant segment extraction as well as summary generation.}
    \label{fig:introduction}
\end{figure}

\begin{figure*}[t]
    \centering
    \includegraphics[width=1\linewidth]{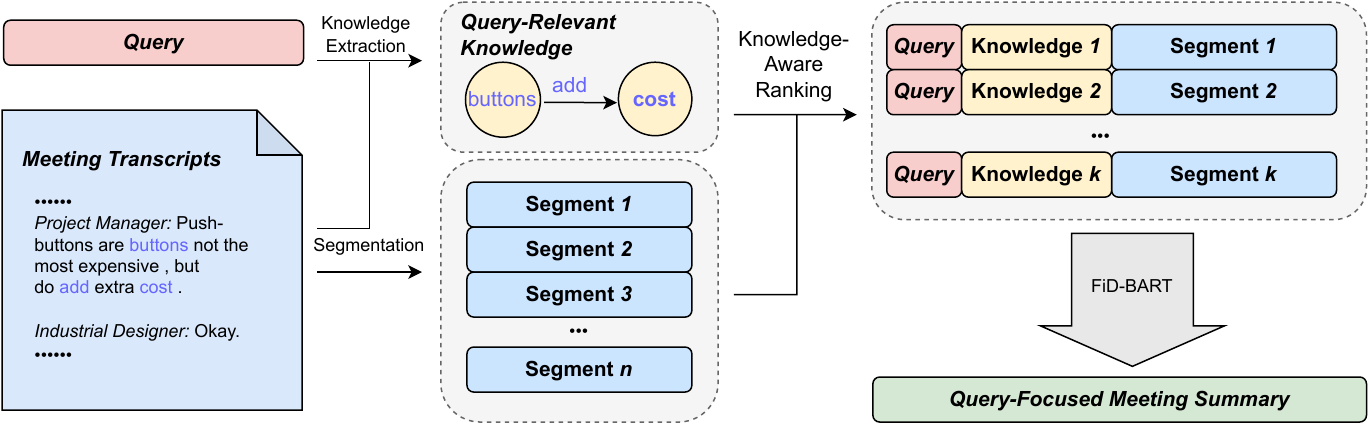}
    \caption{An overview of our proposed framework. We first extract query-relevant knowledge triples based on the query and meeting transcripts and select top-$k$ segments through knowledge-aware ranking. Then we generate the query-focused meeting summary by FiD-BART from query, knowledge and selected meeting transcripts.}
    \label{fig:model}
\end{figure*}

Our contributions in this work are threefold: (1) Our work demonstrates the effectiveness of leveraging query-relevant knowledge in QFMS. (2) We propose KAS, a two-stage framework incorporating query-relevant knowledge for QFMS. (3) Experimental results show that our approach achieves state-of-the-art performance on a QFMS dataset (QMSum). Further analysis and human evaluation indicate the advantage of our method.

\section{Related Works}
Existing QFMS methods can be divided into two categories~\cite{vig2022exploring}: two-stage approaches and end-to-end approaches. Two-stage approaches~\cite{zhong2021qmsum,vig2022exploring,zhang2022summn} first extract query-relevant snippets and then generate the summary upon the extracted snippets, while end-to-end approaches~\cite{zhu2020hierarchical,zhong2022dialoglm,pagnoni2022socratic} directly generate summaries from the whole meeting transcripts. However, both types of methods have their disadvantages. For example, most of the two-stage approaches select query-relevant content based on utterance, which ignores the contextual information between multiple utterances. As for the end-to-end approaches, the computational and memory requirements will increase rapidly when the input text becomes longer, making models challenging to adapt to long meeting transcripts. To our knowledge, we are the first to incorporate query-relevant knowledge in QFMS and demonstrate its effectiveness. Besides, we include multiple utterances in each segment to preserve the contextual information, and our approach can easily extend to process long input text.



\begin{table*}[t]
\centering
\begin{adjustbox}{width=0.8\linewidth}
    \begin{tabular}{lcccc}
        \toprule
        Model                                       & ROUGE-1 & ROUGE-2 & ROUGE-L & Average\\ \midrule
        \multicolumn{3}{l}{\qquad \textit{Two-stage}}  \\
        MaRGE~\cite{xu2021generating}               &31.99  &8.97  &27.93  &22.96 \\
        DYLE~\cite{mao2022dyle}                     &34.42	&9.71  &30.10  &24.74 \\ 
        SUMM$^{N}$~\cite{zhang2022summn}            &34.03  &9.28  &29.48  &24.26 \\ 
        RelReg~\cite{vig2022exploring}              &34.91  &11.91 &30.73  &25.85 \\ 
        RelReg-W~\cite{vig2022exploring}            &36.45  &12.81 &32.28  &27.18 \\ \midrule
        \multicolumn{3}{l}{\qquad \textit{End-to-end}}  \\
        LED~\cite{beltagy2020longformer}            &31.60  &7.80  &20.50  &19.97 \\
        DialogLM~\cite{zhong2022dialoglm}           &34.50	&9.92  &30.27  &24.90 \\
        SegEnc-W~\cite{vig2022exploring}            &37.80  &13.43 &33.38  &28.20 \\
        BART-LS~\cite{xiong2022adapting}            &37.90  &12.10 &33.10  &27.70 \\
        SOCRATIC~\cite{pagnoni2022socratic}         &38.06  &13.74 &33.51  &28.44 \\ \midrule
        KAS (Ours)                                  &\textbf{38.80}  &\textbf{14.01} &\textbf{34.26}  &\textbf{29.02} \\ \bottomrule
    \end{tabular}
    \end{adjustbox}
    \caption{Results on QMSum test set. The previous works can be divided into two-stage and end-to-end approaches.}
    \label{tab:main_results}
\end{table*}
 
\section{Methodology}
This section presents our two-stage framework. Firstly, we introduce the extractor, which extracts query-relevant segments and knowledge from the source document. Then, a generative model synthesizes the query, extracted segments, and knowledge into the final summary. 

\subsection{Knowledge-aware Extractor}
\paragraph{Meeting Transcripts Segmentation}
To preserve the contextual information between utterances, we split the meeting transcripts into segments, and each segment could contain multiple utterances. To do so, given an input meeting transcripts $T$, we will separate it into $n$ segments $S=\{S_{1},S_{2}, ..., S_{n}\}$ where each $S_{i}$ is fewer than $l$ tokens. Specifically, we feed meeting transcripts to the segment utterance by utterance until it reaches $l$.


\paragraph{Knowledge-aware Ranking}
The knowledge-aware ranking approach will select top-$k$ segments according to a combination of semantic search scores and knowledge-aware scores. We apply Multi-QA MPNet~\cite{song2020mpnet} to calculate semantic search scores. Multi-QA MPNet is trained on 215M question-answer pairs from various sources and domains, including Stack Exchange, MS MARCO~\cite{nguyen2016ms}, WikiAnswers~\cite{fader2014open} and many more. Given the query and the segments, the model outputs 768-dimension vectors to represent them. The semantic search score for each segment is computed according to the cosine similarity:

\begin{align}
    Score_{se} = \frac{MPNet(Q) \cdot MPNet(S_{i})}{\left \| MPNet(Q)  \right \| \left \| MPNet(S_{i}) \right \|}
    \label{eq:similarity_function}
\end{align}

To compute the knowledge-aware score for each segment, we first use OpenIE~\cite{angeli2015leveraging} to extractive knowledge triples. Then, to filter out the triples irrelevant to the query, we only keep the triples that contain overlapping words with the query. The knowledge-aware score is obtained by L2 normalizing the number of the remaining triples $m_{i}$ in each segment:
\begin{align}
    Score_{ka} = \frac{m_{i}}{\sqrt{\sum_{i=1}^{n}{|m_{i}|^{2}}}}
    \label{eq:l2_norm}
\end{align}

Finally, we calculate the ranking score by summing the semantic search score and the knowledge-aware score:

\begin{align}
    Score_{rank} = Score_{se} + Score_{ka}
    \label{eq:ranking_score}
\end{align}

The segment with the top-$k$ ranking score will be selected for the next stage of summary generation. We denote the remaining segments as $S=\{S_{1},S_{2}, ..., S_{k}\}$.

\subsection{Generator}
We choose BART-large~\cite{lewis2020bart}, a transformer-based~\cite{vaswani2017attention} generative pre-trained language model, as our backbone model for the generator because of its remarkable performance on text summarization benchmarks. Following the idea of Fusion-in-Decoder (FiD) and its applications in generation tasks~\cite{izacard2021leveraging,su2022read,vig2022exploring}, we employ FiD-BART for encoding multiple segments independently in the encoder and fuse information from all segments in the decoder jointly through the encoder-decoder attention.

To incorporate the extracted knowledge in the summary generation, we use the knowledge as extra inputs other than the query and segments. In detail, we remove the stop words in the knowledge triples and then merge all the remaining words of knowledge triples in each segment as a set of knowledge phrases. The concatenation of each segment $S_{i}$, knowledge phrases $K_{i}$ and the query $Q$ will be processed by FiD-BART encoder:

\begin{align}
    h_{enc}^{i} = Encoder(Q \oplus K_{i} \oplus S_{i})
    \label{eq:encoding}
\end{align}

Finally, the decoder performs encoder-decoder attention over the concatenation of all segments' encoder outputs. In this way, the computational complexity grows linearly with the number of segments rather than quadratically, while jointly processing all segments in the decoder enables the model to aggregate information from multiple segments.

\section{Experiments}
\subsection{Experimental Setup}
We choose the top 12 text segments in the segment selection, with each fewer than 512 tokens. For the summary generator, inspired by the effectiveness of pre-finetuning models on relevant datasets to transfer task-related knowledge in the abstractive summarization~\cite{yu2021adaptsum,vig2022exploring}, we initialize our BART-large model using the checkpoint~\footnote{\url{https://github.com/salesforce/query-focused-sum}} pre-finetuned on WikiSum~\cite{liu2018generating}. See Appendix~\ref{sec:appendix_Experimental} and~\ref{sec:appendix_Baselines} for more details of the experimental setup and baselines.

We evaluate the models on the QMSum~\cite{zhong2021qmsum} dataset, which consists of 1,808 query-summary pairs over 232 meetings from product design, academic, and political committee domains. We report ROUGE~\footnote{\url{https://github.com/pltrdy/files2rouge}}~\cite{lin2004rouge} as the automatic evaluation results.

\begin{table}[t]
\centering
\begin{adjustbox}{width=1\linewidth}
    \begin{tabular}{lcccc}
        \toprule
        model                       & ROUGE-1 & ROUGE-2 & ROUGE-L\\ \midrule
        KAS ($k$=4)               &36.46  &12.06  &31.75 \\
        KAS ($k$=8)               &37.86  &13.75  &33.30 \\
        KAS ($k$=12)              &\textbf{38.80}  &\textbf{14.01}  &\textbf{34.26} \\\bottomrule
    \end{tabular}
    \end{adjustbox}
    \caption{Results Our KAS on the QMSum test set with differen number of $k$. $k$ denotes the number of segments that are selected in the first stage.}
    \label{tab:compression_rate}
\end{table}

\begin{table}[t]
\centering
\begin{adjustbox}{width=1\linewidth}
    \begin{tabular}{lccccc}
        \toprule
        model                  & R-1 & R-2 & R-L & Entity F-1\\ \midrule
        KAS                &\textbf{38.80}  &\textbf{14.01}  &\textbf{34.26} &\textbf{35.59} \\
        w/o KA in ranking      &37.71  &12.96  &32.94 &34.03 \\ 
        w/o KA in generator    &37.52  &13.57  &33.29 &35.06 \\\bottomrule
    \end{tabular}
    \end{adjustbox}
    \caption{Ablation results on the QMSum test set without knowledge-Aware (KA). Our KA modules improve the performance of KAS on both stages.}
    \label{tab:ablation}
\end{table}

\subsection{Main Results}
We compare our proposed model with strong baselines and previous state-of-the-art models. As shown in Table~\ref{tab:main_results}, our approach outperforms both the two-stage and end-to-end methods by a large margin on all evaluation metrics. We also investigate how the number of selected segments in the first stage affects the final performance. As shown in Table~\ref{tab:compression_rate}, when the number of selected segments grows, the quality of the final summary also improves. Previous end-to-end approaches usually outperform two-stage approaches because they encode the entire meeting transcripts, which conduct lots of computing resources. Our approach performs better by encoding 12 segments of input (6,144 tokens) in the second stage generation than the state-of-the-art end-to-end approach that directly encodes 16384 tokens. Therefore, we strike a balance between performance and efficiency, which is essential in real-world applications. Besides, our approach can easily extend to processing long input text since the ranking method is unsupervised.

\subsection{Ablation Study}
We conduct an ablation study to investigate the contribution of the knowledge-aware modules by removing the knowledge-based scoring in the ranking (w/o KA in ranking) and the knowledge input in the summary generation (w/o KA in generator), respectively. Besides, we evaluate the entity-level factual consistency~\cite{nan2021entity} of the summary to test the effectiveness of our knowledge-aware modules in keeping the knowledge entities in generated summaries. Specifically, we report the F-1 scores of the entity overlap (Entity F-1) between the source and the generated summary. Table~\ref{tab:ablation} presents that the model's performance decreases in both metrics, especially the Entity F-1, without the knowledge-aware module, which indicates the effectiveness of our method.

\begin{table}[t]
\centering
\begin{adjustbox}{width=1\linewidth}
    \begin{tabular}{lcccc}
        \toprule
        model         & Fluency  & Relevance & Factual Correctness \\ \midrule
        SegEnc-W      &4.68          &4.60       &3.76 \\ 
        KAS           &4.79          &4.68       &4.47  \\
        Ground truth  &\textbf{4.92}          &\textbf{4.82}       &\textbf{4.60} \\\bottomrule
    \end{tabular}
    \end{adjustbox}
    \caption{Human evaluation results on QMSum test set. }
    \label{tab:human_evaluation}
\end{table}


\subsection{Human Evaluation}
We further conduct a human evaluation to assess the models on fluency, relevance, and factual correctness. We randomly select 50 samples from the QMSum test set and ask three annotators to score the summary from one to five, with higher scores being better. We compare SegEnc-W because it is the best publicly available model. Table~\ref{tab:human_evaluation} illustrates our approach slightly outperforms SegEnc-W according to fluency and relevance and achieves a significantly better score than SegEnc-W on factual correctness with p-value$<$0.05. Since SegEnc-W and our methods both use BART based generators, it is in line with our expectation that we reach similar fluency and relevance scores. Meanwhile, our proposed query-relevant knowledge can help reduce hallucination and generate more factual correctness summaries. More details of the human evaluation setup are in Appendix~\ref{sec:appendix_human_evaluation}.

\section{Conclusion}
In this paper, we propose a two-stage framework named KAS that incorporates query-relevant knowledge in both stages. Extensive experimental results on the QMSum dataset show the effectiveness of our method. We further conduct detailed analysis and human evaluation to prove our method's capacity to generate fluency, relevant and faithful summaries.

\section{Limitations}
Our method is trained and tested on the only publicly available QFMS dataset named QMSum. QMSum consists of three different domains (Academic, Committee and Product), which makes the evaluation robust. However, QMSum only contains 1,808 samples, which is relatively small. We hope larger QFMS datasets be proposed to accelerate the development of this field.

In the first stage of our approach, we extract a fixed length (6,144 tokens) of the meeting transcripts as the second stage input text. Therefore, the model's performance could be affected since some query-relevant information could be cut off in the first stage.

\section{Ethics Statement}
Although our approach can generate query-focused meeting summaries and achieve a much better factual correctness score than other models, we can not avoid generating hallucinated content from the generative models. We recommend that when our approach is deployed in real-world applications, additional post-processing should be carried out to remove unreliable summaries. In addition, we will indicate that the summary we provide is for reference only, and users need to check the original meeting transcripts to get accurate information.

\bibliography{anthology,custom}

\begin{thebibliography}{23}
\expandafter\ifx\csname natexlab\endcsname\relax\def\natexlab#1{#1}\fi

\bibitem[{Angeli et~al.(2015)Angeli, Premkumar, and
  Manning}]{angeli2015leveraging}
Gabor Angeli, Melvin Jose~Johnson Premkumar, and Christopher~D Manning. 2015.
\newblock Leveraging linguistic structure for open domain information
  extraction.
\newblock In \emph{Proceedings of the 53rd Annual Meeting of the Association
  for Computational Linguistics and the 7th International Joint Conference on
  Natural Language Processing (Volume 1: Long Papers)}, pages 344--354.

\bibitem[{Beltagy et~al.(2020)Beltagy, Peters, and
  Cohan}]{beltagy2020longformer}
Iz~Beltagy, Matthew~E Peters, and Arman Cohan. 2020.
\newblock Longformer: The long-document transformer.
\newblock \emph{arXiv preprint arXiv:2004.05150}.

\bibitem[{Fader et~al.(2014)Fader, Zettlemoyer, and Etzioni}]{fader2014open}
Anthony Fader, Luke Zettlemoyer, and Oren Etzioni. 2014.
\newblock Open question answering over curated and extracted knowledge bases.
\newblock In \emph{Proceedings of the 20th ACM SIGKDD international conference
  on Knowledge discovery and data mining}, pages 1156--1165.

\bibitem[{Izacard and Grave(2021)}]{izacard2021leveraging}
Gautier Izacard and {\'E}douard Grave. 2021.
\newblock Leveraging passage retrieval with generative models for open domain
  question answering.
\newblock In \emph{Proceedings of the 16th Conference of the European Chapter
  of the Association for Computational Linguistics: Main Volume}, pages
  874--880.

\bibitem[{Kingma and Ba(2014)}]{kingma2014adam}
Diederik~P Kingma and Jimmy Ba. 2014.
\newblock Adam: A method for stochastic optimization.
\newblock \emph{arXiv preprint arXiv:1412.6980}.

\bibitem[{Lewis et~al.(2020)Lewis, Liu, Goyal, Ghazvininejad, Mohamed, Levy,
  Stoyanov, and Zettlemoyer}]{lewis2020bart}
Mike Lewis, Yinhan Liu, Naman Goyal, Marjan Ghazvininejad, Abdelrahman Mohamed,
  Omer Levy, Veselin Stoyanov, and Luke Zettlemoyer. 2020.
\newblock Bart: Denoising sequence-to-sequence pre-training for natural
  language generation, translation, and comprehension.
\newblock In \emph{Proceedings of the 58th Annual Meeting of the Association
  for Computational Linguistics}, pages 7871--7880.

\bibitem[{Lin(2004)}]{lin2004rouge}
Chin-Yew Lin. 2004.
\newblock Rouge: A package for automatic evaluation of summaries.
\newblock In \emph{Text summarization branches out}, pages 74--81.

\bibitem[{Liu et~al.(2018)Liu, Saleh, Pot, Goodrich, Sepassi, Kaiser, and
  Shazeer}]{liu2018generating}
Peter~J Liu, Mohammad Saleh, Etienne Pot, Ben Goodrich, Ryan Sepassi, Lukasz
  Kaiser, and Noam Shazeer. 2018.
\newblock Generating wikipedia by summarizing long sequences.
\newblock \emph{arXiv preprint arXiv:1801.10198}.

\bibitem[{Mao et~al.(2022)Mao, Wu, Ni, Zhang, Zhang, Yu, Deb, Zhu, Awadallah,
  and Radev}]{mao2022dyle}
Ziming Mao, Chen~Henry Wu, Ansong Ni, Yusen Zhang, Rui Zhang, Tao Yu,
  Budhaditya Deb, Chenguang Zhu, Ahmed Awadallah, and Dragomir Radev. 2022.
\newblock Dyle: Dynamic latent extraction for abstractive long-input
  summarization.
\newblock In \emph{Proceedings of the 60th Annual Meeting of the Association
  for Computational Linguistics (Volume 1: Long Papers)}, pages 1687--1698.

\bibitem[{Nan et~al.(2021)Nan, Nallapati, Wang, dos Santos, Zhu, Zhang,
  Mckeown, and Xiang}]{nan2021entity}
Feng Nan, Ramesh Nallapati, Zhiguo Wang, Cicero dos Santos, Henghui Zhu, Dejiao
  Zhang, Kathleen Mckeown, and Bing Xiang. 2021.
\newblock Entity-level factual consistency of abstractive text summarization.
\newblock In \emph{Proceedings of the 16th Conference of the European Chapter
  of the Association for Computational Linguistics: Main Volume}, pages
  2727--2733.

\bibitem[{Nguyen et~al.(2016)Nguyen, Rosenberg, Song, Gao, Tiwary, Majumder,
  and Deng}]{nguyen2016ms}
Tri Nguyen, Mir Rosenberg, Xia Song, Jianfeng Gao, Saurabh Tiwary, Rangan
  Majumder, and Li~Deng. 2016.
\newblock Ms marco: A human generated machine reading comprehension dataset.
\newblock \emph{choice}, 2640:660.

\bibitem[{Pagnoni et~al.(2022)Pagnoni, Fabbri, Kry{\'s}ci{\'n}ski, and
  Wu}]{pagnoni2022socratic}
Artidoro Pagnoni, Alexander~R Fabbri, Wojciech Kry{\'s}ci{\'n}ski, and
  Chien-Sheng Wu. 2022.
\newblock Socratic pretraining: Question-driven pretraining for controllable
  summarization.
\newblock \emph{arXiv preprint arXiv:2212.10449}.

\bibitem[{Song et~al.(2020)Song, Tan, Qin, Lu, and Liu}]{song2020mpnet}
Kaitao Song, Xu~Tan, Tao Qin, Jianfeng Lu, and Tie-Yan Liu. 2020.
\newblock Mpnet: Masked and permuted pre-training for language understanding.
\newblock \emph{Advances in Neural Information Processing Systems},
  33:16857--16867.

\bibitem[{Su et~al.(2022)Su, Li, Zhang, Shang, Jiang, Liu, and
  Fung}]{su2022read}
Dan Su, Xiaoguang Li, Jindi Zhang, Lifeng Shang, Xin Jiang, Qun Liu, and
  Pascale Fung. 2022.
\newblock Read before generate! faithful long form question answering with
  machine reading.
\newblock In \emph{Findings of the Association for Computational Linguistics:
  ACL 2022}, pages 744--756.

\bibitem[{Vaswani et~al.(2017)Vaswani, Shazeer, Parmar, Uszkoreit, Jones,
  Gomez, Kaiser, and Polosukhin}]{vaswani2017attention}
Ashish Vaswani, Noam Shazeer, Niki Parmar, Jakob Uszkoreit, Llion Jones,
  Aidan~N Gomez, {\L}ukasz Kaiser, and Illia Polosukhin. 2017.
\newblock Attention is all you need.
\newblock \emph{Advances in neural information processing systems}, 30.

\bibitem[{Vig et~al.(2022)Vig, Fabbri, Kry{\'s}ci{\'n}ski, Wu, and
  Liu}]{vig2022exploring}
Jesse Vig, Alexander~Richard Fabbri, Wojciech Kry{\'s}ci{\'n}ski, Chien-Sheng
  Wu, and Wenhao Liu. 2022.
\newblock Exploring neural models for query-focused summarization.
\newblock In \emph{Findings of the Association for Computational Linguistics:
  NAACL 2022}, pages 1455--1468.

\bibitem[{Xiong et~al.(2022)Xiong, Gupta, Toshniwal, Mehdad, and
  Yih}]{xiong2022adapting}
Wenhan Xiong, Anchit Gupta, Shubham Toshniwal, Yashar Mehdad, and Wen-tau Yih.
  2022.
\newblock Adapting pretrained text-to-text models for long text sequences.
\newblock \emph{arXiv preprint arXiv:2209.10052}.

\bibitem[{Xu and Lapata(2021)}]{xu2021generating}
Yumo Xu and Mirella Lapata. 2021.
\newblock Generating query focused summaries from query-free resources.
\newblock In \emph{Proceedings of the 59th Annual Meeting of the Association
  for Computational Linguistics and the 11th International Joint Conference on
  Natural Language Processing (Volume 1: Long Papers)}, pages 6096--6109.

\bibitem[{Yu et~al.(2021)Yu, Liu, and Fung}]{yu2021adaptsum}
Tiezheng Yu, Zihan Liu, and Pascale Fung. 2021.
\newblock Adaptsum: Towards low-resource domain adaptation for abstractive
  summarization.
\newblock In \emph{Proceedings of the 2021 Conference of the North American
  Chapter of the Association for Computational Linguistics: Human Language
  Technologies}, pages 5892--5904.

\bibitem[{Zhang et~al.(2022)Zhang, Ni, Mao, Wu, Zhu, Deb, Awadallah, Radev, and
  Zhang}]{zhang2022summn}
Yusen Zhang, Ansong Ni, Ziming Mao, Chen~Henry Wu, Chenguang Zhu, Budhaditya
  Deb, Ahmed Awadallah, Dragomir Radev, and Rui Zhang. 2022.
\newblock Summn: A multi-stage summarization framework for long input dialogues
  and documents: A multi-stage summarization framework for long input dialogues
  and documents.
\newblock In \emph{Proceedings of the 60th Annual Meeting of the Association
  for Computational Linguistics (Volume 1: Long Papers)}, pages 1592--1604.

\bibitem[{Zhong et~al.(2022)Zhong, Liu, Xu, Zhu, and Zeng}]{zhong2022dialoglm}
Ming Zhong, Yang Liu, Yichong Xu, Chenguang Zhu, and Michael Zeng. 2022.
\newblock Dialoglm: Pre-trained model for long dialogue understanding and
  summarization.
\newblock In \emph{Proceedings of the AAAI Conference on Artificial
  Intelligence}, volume~36, pages 11765--11773.

\bibitem[{Zhong et~al.(2021)Zhong, Yin, Yu, Zaidi, Mutuma, Jha, Hassan,
  Celikyilmaz, Liu, Qiu et~al.}]{zhong2021qmsum}
Ming Zhong, Da~Yin, Tao Yu, Ahmad Zaidi, Mutethia Mutuma, Rahul Jha, Ahmed
  Hassan, Asli Celikyilmaz, Yang Liu, Xipeng Qiu, et~al. 2021.
\newblock Qmsum: A new benchmark for query-based multi-domain meeting
  summarization.
\newblock In \emph{Proceedings of the 2021 Conference of the North American
  Chapter of the Association for Computational Linguistics: Human Language
  Technologies}, pages 5905--5921.

\bibitem[{Zhu et~al.(2020)Zhu, Xu, Zeng, and Huang}]{zhu2020hierarchical}
Chenguang Zhu, Ruochen Xu, Michael Zeng, and Xuedong Huang. 2020.
\newblock A hierarchical network for abstractive meeting summarization with
  cross-domain pretraining.
\newblock In \emph{Findings of the Association for Computational Linguistics:
  EMNLP 2020}, pages 194--203.

\end{thebibliography}
\bibliographystyle{acl_natbib}

\appendix
\section{Experimental Details}
\label{sec:appendix_Experimental}
We use the off-the-shelf model Multi-QA MPNet~\footnote{\url{https://huggingface.co/sentence-transformers/multi-qa-mpnet-base-cos-v1}} for semantic searching. We initialize the generator from the off-the-shelf BART-large model through Hugging Face. We use learning rates $6e^{-5}$ following~\cite{lewis2020bart} and Adam optimizer~\cite{kingma2014adam} to fine-tune KAS. We use a batch size of one and train all models on one RTX 3090 for ten epochs. During decoding, we use beam search with a beam size of five and decode until an end-of-sequence token is emitted. The final results are the average test set performance on the best three checkpoints in the validation set. In addition, the QMSum dataset has MIT License, so we can freely use it.

\section{Baselines Details}
\label{sec:appendix_Baselines}
We conduct both two-stage and end-to-end baselines for comparison. MaRGE~\cite{xu2021generating} used \textbf{Ma}sked \textbf{R}OU\textbf{GE} extractor in the first stage. DYLE~\cite{mao2022dyle} is a dynamic latent extraction approach for abstractive long-input summarization. SUMM$^{N}$~\cite{zhang2022summn} is a multi-stage summarization framework for long input dialogues and documents. RelReg (RELevance REGression)~\cite{vig2022exploring} is similar to MaRGE, which trains a relevance prediction model directly on QFS data using the original, non-masked query. RelReg-W~\cite{vig2022exploring} uses the same framework as RelReg but initializes the generator from the WikiSum pre-trained checkpoint. LED~\cite{beltagy2020longformer} is an encoder-decoder transformer-based model that employs efficient attention to process long input text. DialogLM~\cite{zhong2022dialoglm} is a pre-trained neural encoder-decoder model for long dialogue understanding and summarization, and it uses a hybrid attention approach by combining sparse local attention with dense global attention. SegEnc-W~\cite{vig2022exploring} is a fusion-in-decoder method that initialized the model from WikiSum~\cite{liu2018generating}. BART-LS~\cite{xiong2022adapting} used pooling-augmented blockwise attention to improve the efficiency and pre-trains the model with a masked-span prediction task with spans of varying lengths. SOCRATIC~\cite{pagnoni2022socratic} uses a question-driven, unsupervised pretraining objective to improve controllability in summarization tasks, and it follows the SegEnc framework.

\section{Human Evaluation}
\label{sec:appendix_human_evaluation}
In Table~\ref{tab:human_evaluation}, we conduct a human evaluation of the fluency, relevancy and factual correctness of the generated summaries from the QMSum dataset. In detail, we randomly sample 50 summaries from KAS, SegEnc-W and the ground truth summary from the same meeting transcripts for comparison. Assessments are scored on a scale of one to five, with higher scores being better. Fluency means the summary is grammatically correct and contextually coherent. Relevancy means the summary is relevant to the query. Factual correctness means the summary is faithful according to the information given in the meeting transcripts and does not contain hallucinations. We assign each summary to three annotators and take the average score as the final result. In total, we used three annotators from Hong Kong, and all annotators voluntarily participated in the human evaluation. All annotators agree to have their evaluation results included as part of this paper's results.

\end{document}